\pdfoutput=1

\documentclass[11pt]{article}

\usepackage[preprint]{acl}

\usepackage{times}
\usepackage{latexsym}

\usepackage[T1]{fontenc}

\usepackage[utf8]{inputenc}

\usepackage{microtype}

\usepackage{inconsolata}

\usepackage{graphicx}

\usepackage{subcaption}
\usepackage{booktabs}
\usepackage{amsmath}
\usepackage{multirow}

\title{Flaming-hot Initiation with Regular Execution Sampling for Large Language Models}

\author{Weizhe Chen \\
	University of Southern California \\
	\texttt{weizhech@usc.edu} \\\And
	Zhicheng Zhang \\
	Carnegie Mellon University \\
	\texttt{zhichen3@cs.cmu.edu} \\\AND
	Guanlin Liu, 
	Renjie Zheng,
	Wenlei Shi,
	Chen Dun,
	Zheng Wu,
	Xing Jin, 
	Lin Yan \\
	ByteDance Inc.\\
	\texttt{\{guanlin.liu, renjie.zheng, wenlei.shi\}@bytedance.com} \\
	\texttt{\{chen.dun, zheng.wu1, jinxing.9, neil\}@bytedance.com}\\
}

\begin{document}
	\maketitle
	\begin{abstract}
		Since the release of ChatGPT, large language models (LLMs) have demonstrated remarkable capabilities across various domains. 
		A key challenge in developing these general capabilities is efficiently sourcing diverse, high-quality data. This becomes especially critical in reasoning-related tasks with sandbox checkers, such as math or code, where the goal is to generate correct solutions to specific problems with higher probability.
		In this work, we introduce Flaming-hot Initiation with Regular Execution (FIRE) sampling, a simple yet highly effective method to efficiently find good responses. Our empirical findings show that FIRE sampling enhances inference-time generation quality and also benefits training in the alignment stage. Furthermore, we explore how FIRE sampling improves performance by promoting diversity and analyze the impact of employing FIRE at different positions within a response.
	\end{abstract}
	
	\section{Introduction}
	
	Large language models (LLMs) have achieved remarkable success in a wide range of tasks since the release of ChatGPT~\citep{chatgpt}.
	In addition to traditional natural language processing tasks such as summarization and sentiment analysis,
	LLMs have demonstrated effectiveness in many new domains, including
	code generation~\citep{chen2023teaching, roziere2023code}, human-computer interaction~\citep{li2023camel}, and math problem-solving \citep{DBLP:conf/nips/Wei0SBIXCLZ22, yu2023metamath}.
	Although standalone LLMs have limited reasoning capabilities
	\citep{sun2023survey, valmeekam2024planbench, chen2024solving},
	researchers have tried to enhance them by incorporating tool-use and developing integrated systems known as LLM agents~\citep{xi2023rise,wang2024survey}, which expands the applications of LLMs to more general domains like robot control~\citep{wang2023prompt} and autonomous driving~\citep{mao2023agent}.
	
	{To develop general capabilities, LLMs are typically trained through a three-stage process: pretraining, supervised fine-tuning (SFT), and alignment~\citep{bai2022training,ouyang2022training}. During pretraining, the model learns from a vast array of data gathered from publicly available sources. Then, in the SFT and alignment stages, the model's abilities are further refined, allowing it to increase reasoning abilities and better follow users' instructions.}
	In order to enhance reasoning tasks, a sandbox checker — a tool used to verify the correctness of solutions — is often used during training~\cite{liu2023rltf}. Therefore, one of the key challenges in achieving effective and efficient training is determining how to obtain more successful samples within a fixed number of trials, particularly when addressing complex problems.
	
	In this paper, we introduce Flaming-hot Initiation with Regular Execution (FIRE), a simple yet effective sampling method for training large language models. Inspired by recent findings on attention sink~\citep{xiao2023efficient}, our approach begins by sampling the initial token at a very high temperature and proceeds with the regular sampling process for the remaining sequence.
	Our algorithm can be viewed as a simplified and more general version of CoT-decoding~\citep{wang2024chain}, especially with a focus on training in math and coding domains where a sandbox checker is available at a relatively cheap cost.
	
	We first show that our method, at inference time, can improve the pass rate within N trials (pass@n), also known as the best-of-N (BoN) when only the correctness of the final answer is considered. To demonstrate its effectiveness in training, we show that it can be directly integrated into the reinforcement learning process of large language models. Our approach proves to be effective across multiple open-source models and various LLM capabilities, including mathematical reasoning and coding. We highlight how our method promotes diversity in generated samples, a key factor linked to performance improvements in pass rate. Importantly, this diversity is maintained even after training with our sampling method, indicating room for further enhancement. We also discuss the effects of simple variations of our method, where the temperature change occurs mid-process rather than at the start, on performance outcomes.
	
	\section{Related Works}

	Researchers have been exploring two primary directions to efficiently improve response quality under a frozen pre-trained LLM. The first direction focuses on prompting techniques such as Chain-of-Thought \cite{DBLP:conf/nips/Wei0SBIXCLZ22} and Tree-of-Thought \cite{yao2024tree}. The second direction involves letting LLMs fix their own mistakes~\cite{wang2022self, yao2022react, shinn2023reflexion, madaan2024self, chen2024reprompt}.
	In line with these two directions, there has been increasing focus on controlled decoding in LLMs to enhance reasoning capabilities during inference, ranging from search-based approaches applied to policy models~\cite{mudgal2023controlled,huang2024deal} to utilizing value models trained in the alignment phase~\cite{liu2023making, feng2023alphazero}.
	
	In this paper, we also focus on inference time; however, our approach extends to the sampling processes used during the training of large language models, as commonly practiced in InstructGPT~\cite{ouyang2022training}.
	This process consists of three key stages: pretraining, supervised fine-tuning (SFT), and alignment, also known as reinforcement learning with human feedback (RLHF). For large language models trained in this paradigm, there could be some helpful properties that, without strong theoretical guarantees, are empirically true and thus helpful for LLMs. Our work is related to attention sink \cite{xiao2023efficient}. 
	An attention sink refers to a token or set of tokens that disproportionately receive attention from other tokens during the attention mechanism within transformer architectures. 
	In their study, they found that one of the most identifiable tokens was shown to be the initial token. While there are no theoretical guarantees, they propose an intuition that initial tokens are visible and used in all later token generations, making them more readily trained to be attention sinks. 
	
	Our work is closely related to CoT-decoding~\cite{wang2024chain}, which uncovers the CoT-paths by enumerating over the top-k alternative tokens and aggregating the responses by scoring the decoded responses with confidence on the final answer.
	However, our approach differs in three key aspects: (1) we introduce a differentiable sampling method that can be directly integrated with existing inference and training frameworks, (2) we focus on improving model performance in scenarios with a sandbox checker, where aggregating responses is less data-efficient, and (3) our method operates without assumptions about the prompts, even when a chain of thought (CoT) is included, extending beyond the scope of CoT-decoding.
	Prior work has explored dynamic temperature allocation during processing \cite{chang2023kl, zhu2023improving, zhang2024edt}. However, these approaches restrict the temperature values to the range of $[0,1]$, which contrasts sharply with the extreme temperature settings employed in our algorithm.
	
	Orthogonal to our work, numerous efforts in the NLP domain have proposed prompt-based algorithms to enhance the diversity of generated responses \cite{naik2023diversity, huang2023enhancing, wu2023large, zhou2024paraphrase}. In contrast, our method focuses on scenarios where the prompt remains fixed, improving generation diversity by controlling the sampling process without relying on domain-specific modifications to the prompt. Notably, our approach is complementary to these works and can be combined with prompt-based methods to further enhance diversity, particularly in cases where generating a larger number of diverse samples for the same question is required.

	\section{Flaming-hot Initiation Regular Execution}

	\subsection{Method}
	
	In this work, we propose a sampling method, Flaming-hot Initiation with Regular Execution (FIRE), inspired by the attention sink phenomenon~\cite{xiao2023efficient} that demonstrates the importance of initial tokens.
	
	FIRE first samples the initial token at a very high temperature $p \gg 1$, combined with top-$k$ filtering to make the candidate tokens more controllable. At higher temperatures, the candidate tokens are sampled from a probability distribution that approaches uniform sampling. After the initial token is sampled, FIRE proceeds with the decoding stage using a regular temperature setting.
	
	Our approach FIRE is similar to CoT-decoding~\cite{wang2024chain} that enumerates the top-$k$ candidates of the initial token. However, while CoT-decoding focuses more on the decoding stage and extracting Chain-of-Thought without prompt, our approach FIRE serves as a general differentiable sampling method, which can be combined with existing sampling frameworks and can be more efficient in the training stage where a sandbox checker that judges whether a specific answer is correct or not is available with a cheap cost. 
	
	While FIRE can be applied to any token in the decoding stage, we restrict its application to the initial token to prevent the generation of random tokens that are wrong in the context.
		For example, if we apply FIRE after the prefix "1+2=", it would sample, in addition to the token "3", other tokens like "4" or "5", which are very likely to be wrong.
		In contrast, since FIRE is only applied to the initial token, it would unlikely lead to broken sentences or code with syntax errors. In our empirical experiments, we found that the initial token frequently consists of words like "Let's", "Sure", "So", and "The", which do not directly convey any information.
		But what these initial tokens affect is the reasoning steps afterward, with the same intuition as StreamingLLM~\cite{xiao2023efficient}.

		\subsection{Experiments}
		
		In this section, we evaluate our algorithm, FIRE, by addressing several key research questions that guide our experiments.

	\paragraph{How effective is FIRE during inference?}
	
	\begin{table}[]
		\small
			\begin{tabular}{@{}clcccc@{}}
				\toprule
				&     & \multicolumn{2}{c}{Regular} & \multicolumn{2}{c}{FIRE} \\
				& Model         & Pass\%       & \#EA     & Pass\%    & \#EA    \\ \midrule
				& DeepSeek & 97.57          & 2.26        & \textbf{98.71}        & \textbf{2.76}      \\
				GSM8K & Gemma-2    & 86.81          & 3.87        & \textbf{87.57}        & \textbf{4.01}      \\
				& Qwen2     & 95.90          & 2.58        & \textbf{98.25}        & \textbf{3.17}      \\
				& Qwen2-RL    & 96.90          & 2.63        & \textbf{97.90}        & \textbf{3.26}      \\\midrule
				& DeepSeek & 76.16          & 5.63        & \textbf{78.16 }       & \textbf{7.89}      \\
				MATH  & Gemma-2    & 49.20          & 9.24        & \textbf{51.48}        & \textbf{10.39}\\
				& Qwen2     & 76.60          & 7.44        & \textbf{79.08}        & \textbf{9.03}      \\ 
				& Qwen2.5-72B & 79.30          & 2.39        & \textbf{80.40}        & \textbf{2.60}      \\\bottomrule
			\end{tabular}%
		\caption{Inference results for different models on different datasets with best hyperparameters combinations. Specifically, Qwen2-RL is a fine-tuned model trained by ourselves. We show the pass rate (\%) with 40 samples, and the effective answers (EA) among the 40 samples.}
		\label{table:inference_res_math}
	\end{table}

	\begin{table}[t]
		\small
		\centering
		\begin{tabular}{@{}lcccc@{}}
			\toprule
			& \multicolumn{2}{c}{Regular} & \multicolumn{2}{c}{FIRE} \\
			& Pass@1       & Pass@10       & Pass@1     & Pass@10     \\ \midrule
			MBPP  & \textbf{61.2}         & 82.8          & 50.6       & \textbf{86.6}        \\
			MBPP+ & \textbf{52.7}         & 74.2          & 44.1       & \textbf{77.0}        \\ \bottomrule
		\end{tabular}
		\caption{Pass rate (\%) with different number of samples from Qwen2-7B-Instruct on MBPP and MBPP+. }
		\label{table:inference_res_code}
	\end{table}

	\begin{table}[t]
		\small
		\centering
		\begin{tabular}{ccccccc}
			\toprule
			\multicolumn{3}{c}{} & \multicolumn{2}{c}{Regular} & \multicolumn{2}{c}{FIRE} \\
			p & k & min-p & n=10 & n=40 & n=10 & n=40 \\
			\midrule
			\multirow{4}{*}{0.7} & \multirow{2}{*}{16} & 0.01 & 66.4 & 75.8 & \textbf{70.0} & \textbf{78.9} \\
			&  & 0 & 66.4 & 75.8 & \textbf{70.0} & \textbf{78.9} \\
			\cmidrule{2-7}
			& \multirow{2}{*}{32} & 0.01 & 66.2 & 75.3 & \textbf{70.1} & \textbf{78.9} \\
			&  & 0 & 66.2 & 75.2 & \textbf{70.1} & \textbf{78.9} \\
			\midrule
			\multirow{4}{*}{0.9} & \multirow{2}{*}{16} & 0.01 & 66.1 & 76.6 & \textbf{69.5} & \textbf{78.9} \\
			&  & 0 & 66.2 & 76.6 & \textbf{69.5} & \textbf{78.9} \\
			\cmidrule{2-7}
			& \multirow{2}{*}{32} & 0.01 & 66.7 & 76.4 & \textbf{69.5} & \textbf{79.1} \\
			&  & 0 & 66.8 & 74.4 & \textbf{69.1} & \textbf{79.0} \\
			\bottomrule
		\end{tabular}
		\caption{Pass rate (\%) for Qwen2-7B-Instruct on MATH dataset with different hyperparameter combinations. p: nucleus sampling parameter, k: top-k sampling parameter, min-p: minimum probability threshold (0 indicates min-p is not used). n=$10$ and n=$40$ represent the number of samples for calculating the pass rate.}
		\label{table:inference_hyperparameters}
	\end{table}
	
	\begin{figure*}[t]
		\centering
		\begin{subfigure}[b]{0.45\linewidth}
			\centering
			\includegraphics[width=0.95\columnwidth]{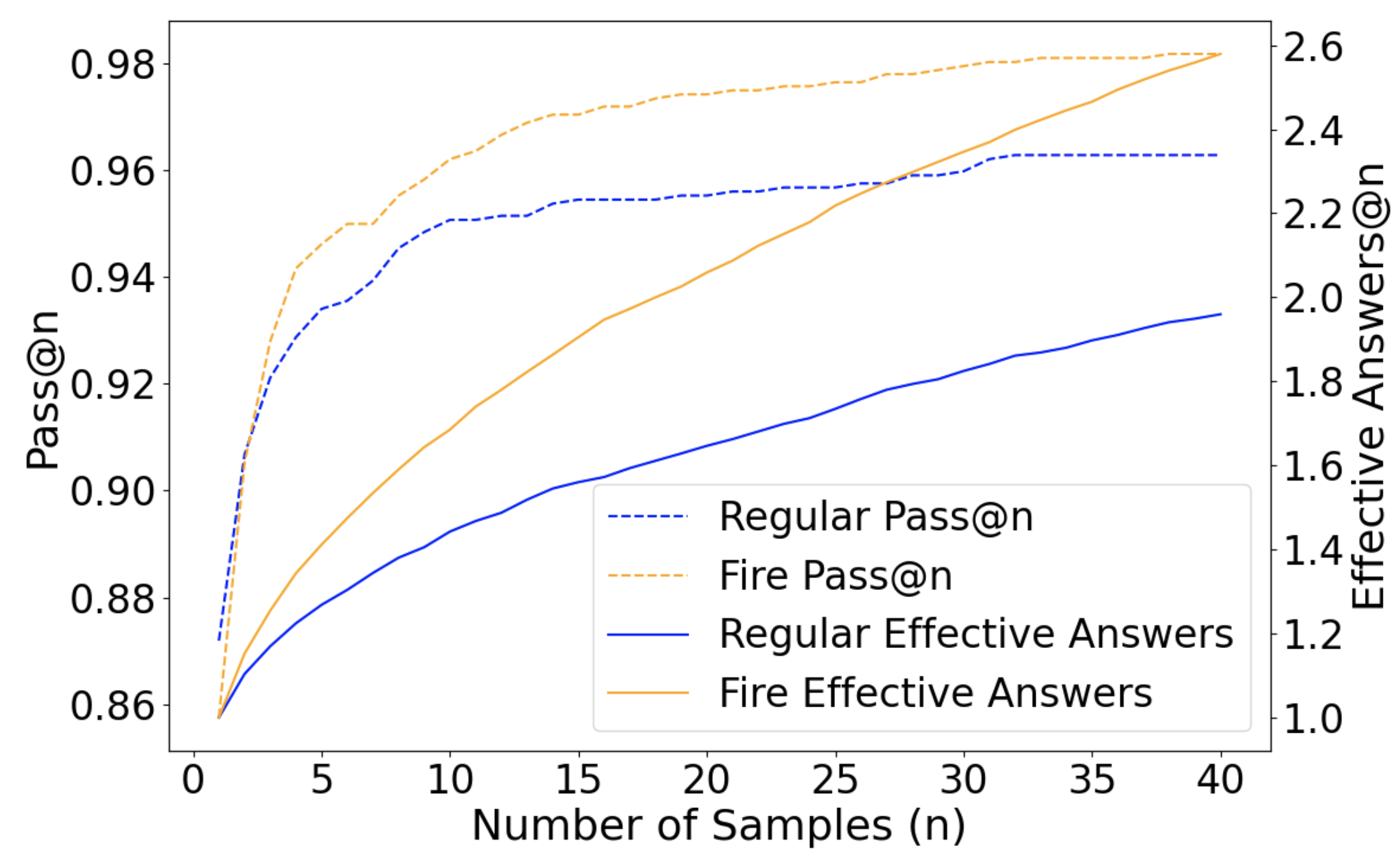}
			\caption{Deepseek-Code-v2-Lite-Instruct}
		\end{subfigure}
		\centering
		\begin{subfigure}[b]{0.45\linewidth}
			\centering
			\includegraphics[width=0.95\columnwidth]{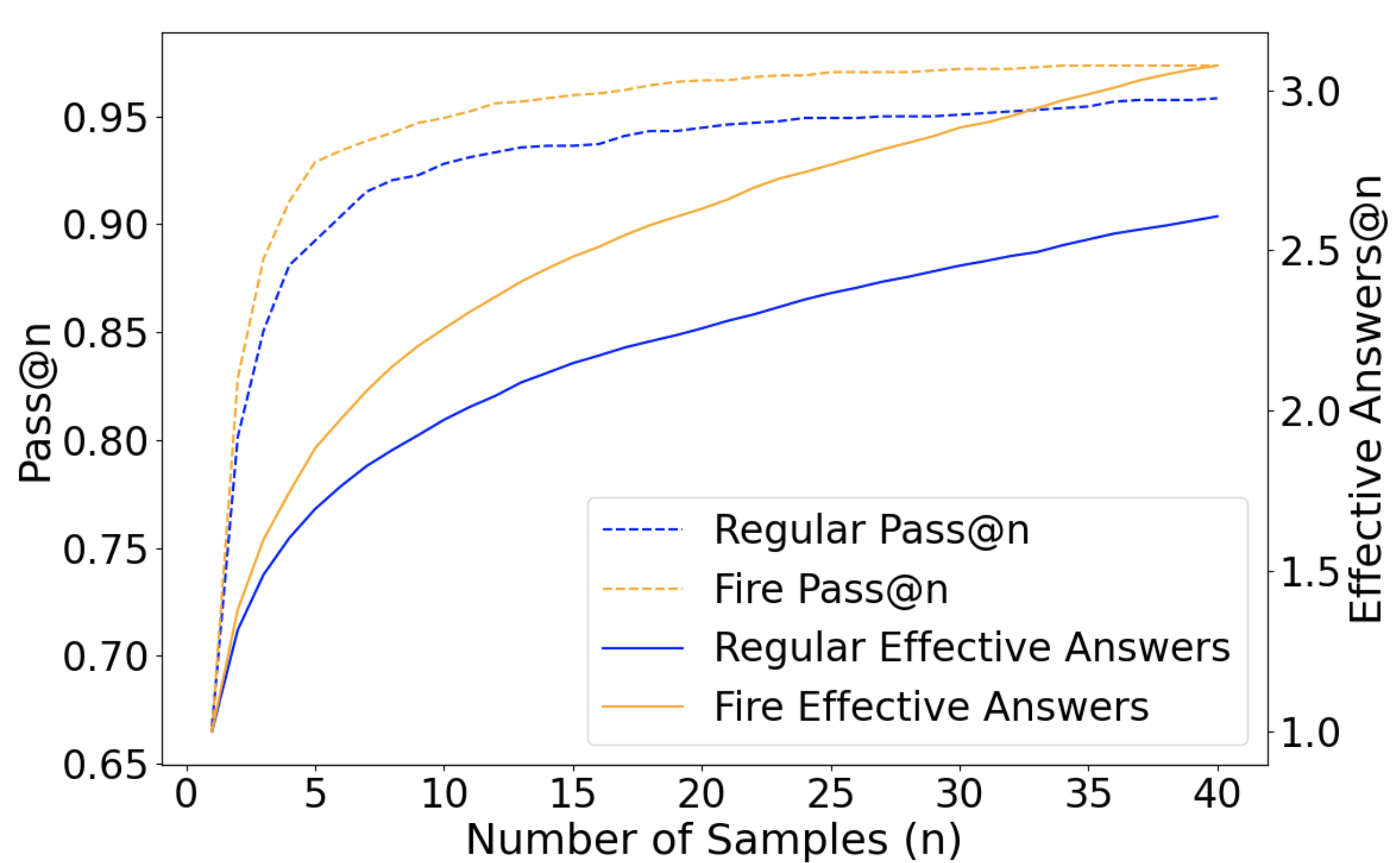}
			\caption{Qwen2-7B-Instruct}
		\end{subfigure}
		
		\caption{Curves for pass rate and number of effective answers with different numbers of samples on GSM8K. }
		\label{fig:bon_curves}
	\end{figure*}
	
	We first showcase the effectiveness of FIRE sampling in inference-only scenarios. 
	We tested four open-source models: 
	Qwen2-7B-Instruct (Qwen2) \cite{yang2024qwen2}, Qwen2.5-72B-Instruct (Qwen2.5-72B)~\cite{yang2024qwen2}, DeepSeek-coder-v2-Instruct (DeepSeek)\cite{zhu2024deepseek}, and Gemma-2-2b-it (Gemma-2)\cite{team2024gemma}, on a diverse set of datasets, including GSM8K \cite{cobbe2021gsm8k}, MATH \cite{hendrycksmath2021}, and MBPP(+) \cite{austin2021program, evalplus}. 
	In GSM8K and MATH, we extend the prompts with phrase "Please reason step-by-step" to ensure CoT reasoning in models' responses, a setting where the original motivation of CoT-decoding becomes less meaningful as CoT paths would naturally occur.
	For the regular sampling settings, we use a combination of nucleus sampling and top-k sampling.
	To ensure a fair comparison, we conducted a thorough enumeration over hyperparameters, including $p$, $k$, and min-p~\cite{gante2023minpsampling}. Table \ref{table:inference_res_math} and Table~\ref{table:inference_res_code} present the aggregated results, where the reported numbers represent the best outcomes from the enumeration. We observe that FIRE consistently improves the pass rate compared to regular settings across all models on different benchmarks. 
	To further demonstrate the consistent improvement over different hyperparameters, we provide an example result of Qwen2-7B-Instruct on the MATH dataset in Table \ref{table:inference_hyperparameters}. Full results for all models and datasets are provided in the appendix. Table \ref{table:inference_hyperparameters} reveals that although FIRE may alter the hyperparameter combination that yields optimal performance, it consistently outperforms regular sampling across all hyperparameter combinations.

	\paragraph{Why is FIRE effective?}
	FIRE introduces more diversity to the initial token that is generated at a high temperature, and due to the strong attention scores towards initial tokens~\cite{xiao2023efficient}, this diversity benefits the entire subsequent generation.
	To measure diversity quantitatively, we use the number of unique answers (effective answers) within a set of responses as our metric. We choose not to use some popular metrics like n-grams since we only control the initial token, and in tasks with long reasoning paths, such as math and coding, similar n-grams will likely always appear, making it unsuitable for measuring diversity.
	As shown in Figure~\ref{fig:bon_curves}, Table~\ref{table:inference_res_math} (\#EA), FIRE demonstrates increased diversity across various models and datasets, which contributes to enhanced pass@n performance. As anticipated, FIRE does not improve Pass@1 performance due to its focus on promoting diversity. However, it consistently delivers improvements when more samples are considered.
	
	\begin{table}[t]
		\small
		\centering
		\begin{tabular}{@{}llcc@{}}
			\toprule
			Dataset & Model    & PPO   & PPO+FIRE \\ \midrule
			& Deepseek & 80.64 & \textbf{82.16}   \\
			GSM8K   & Qwen2     & 80.16 & \textbf{82.02}    \\
			& Gemma    & 40.39 & \textbf{42.91}    \\
			& Gemma-2  & 58.07 & \textbf{61.20}    \\ \midrule
			MATH    & Qwen2     & 53.50 & \textbf{55.07}    \\ \bottomrule
		\end{tabular}
		\caption{Pass@1 on GSM8K and Math for Different models trained with PPO with different sampling.}
		\label{table:rl_res}
	\end{table}
	
	\begin{table}[t]
		\small
		\centering
		\begin{tabular}{@{}lccccc@{}}
			\toprule
			& 1st-line & 2nd-line & 3rd-line   & PRM-line \\ \midrule
			Regular      &   66.40              &   74.36          &        74.77    & 75.73 \\
			FIRE &    \textbf{69.98}           &  \textbf{74.96} & \textbf{75.92}  & \textbf{78.21} \\ \bottomrule
		\end{tabular}
		\caption{Pass@10 Results from Qwen2-7B-Instruct on the training set of MATH dataset for FIRE variants with different sampling points, compared to regular sampling method that does not change the temperature.}
		\label{table:middle_high_temp}
	\end{table}
	
	\begin{table*}[tbh!]
		\small
		\centering
		\begin{tabular}{@{}lllllll@{}}
			\toprule
			Dataset & Model    & Sampling & Pass@1           & Pass@5           & Pass@10          & Pass@20          \\ \midrule
			GSM8K   & Qwen     & Reg      & $66.95 \pm 0.09$ & $89.33 \pm 0.06$ & $92.58 \pm 0.04$ & $94.48 \pm 0.03$ \\
			&          & FIRE     & $65.65 \pm 0.12$ & $92.92 \pm 0.04$ & $95.58 \pm 0.03$ & $97.20 \pm 0.02$ \\
			& Deepseek & Reg      & $87.02 \pm 0.06$ & $93.64 \pm 0.03$ & $95.06 \pm 0.03$ & $96.02 \pm 0.02$ \\
			&          & FIRE     & $85.85 \pm 0.07$ & $94.86 \pm 0.03$ & $96.76 \pm 0.02$ & $97.60 \pm 0.02$ \\
			MATH    & Qwen     & Reg      & $35.94 \pm 0.05$ & $59.32 \pm 0.03$ & $66.04 \pm 0.03$ & $71.19 \pm 0.02$ \\
			&          & FIRE     & $39.64 \pm 0.05$ & $63.34 \pm 0.03$ & $69.75 \pm 0.03$ & $74.77 \pm 0.02$ \\
			& Deepseek & Reg      & $51.03 \pm 0.04$ & $64.45 \pm 0.03$ & $68.58 \pm 0.02$ & $71.80 \pm 0.01$ \\
			&          & FIRE     & $48.79 \pm 0.05$ & $65.39 \pm 0.02$ & $70.25 \pm 0.02$ & $73.99 \pm 0.02$ \\ \bottomrule
		\end{tabular}
		\caption{Average and standard deviation of pass rate results for different models on different datasets. }
		\label{table:statistical}
	\end{table*}
	
	\paragraph{Is FIRE helpful when integrated into training?}
	Having established that our method improves pass@n by improving diversity, we directly apply FIRE to boost language model training. 
	To test this, we use Proximal Policy Optimization (PPO) \cite{schulman2017proximal} to finetune several models using the GSM8K and MATH datasets, and assess their performance through the final pass rate for single samples (Pass@1).
	As shown in Table~\ref{table:rl_res}, integrating FIRE into the training process leads to an improvement in Pass@1.  Notably, even though each data point is sampled only once during PPO training following common practice ~\cite{ouyang2022training, sheng2024hybridflow}, our method still yields improvements. The results also show that the improvements are consistent for different models. Furthermore, after our RL training, the model still exhibits diversity and continues to benefit from inference-time pass rate improvements, as evidenced by Qwen2-RL in Table \ref{table:inference_res_math}. Consequently, FIRE can be applied iteratively to refine the model, leading to an even bigger improvement margin.
	
	\paragraph{{Can FIRE sampling work in mid-sequence?}}
	
	Finally, we explore the effect of applying FIRE sampling midway through a response. We first construct a dataset that ensures the correctness of the initial sentences, 
	by utilizing a Process Reward Model (PRM) to identify the first sentences at which the response becomes incorrect. We then evaluate the effect of applying FIRE sampling at the beginning of different sentences (1st, 2nd, and 3rd-line) or at the first token deemed incorrect by the PRM ("PRM-line"). We refer the reader to the appendix for a more detailed description of the construction of this dataset.
	As shown in Table~\ref{table:middle_high_temp}, while FIRE sampling offers benefits throughout different settings, its advantages diminish for tokens beyond the initial ones, despite an overall increase in accuracy due to the prefix guaranteed to be correct.
	
	\paragraph{Is the improvement stable?}
	
	In previous sections and the appendix, we demonstrated that FIRE consistently enhances performance across various models, datasets, and hyperparameter configurations. Here, we further substantiate these findings by evaluating their statistical significance. Specifically, for each question, we leverage a pool of 40 pre-collected samples and randomly select different subsets of 
	$n$ samples, repeating this process 10 times. The results, summarized in Table~\ref{table:statistical}, reveal minimal standard deviations across the datasets, underscoring the robustness and statistical stability of the improvements.
	\section{Conclusion}
	
	In this paper, we introduced a novel sampling method called Flaming-hot Initiation with Regular Execution (FIRE). Through empirical analysis, we demonstrated that FIRE enhances both inference-time performance and reinforcement learning, particularly when a chain of thought is integrated into the prompt. We showed that FIRE improves generation diversity, and we believe that this diversity contributes to its overall effectiveness. Additionally, we explored several variants of FIRE that modify the sampling process not only immediately after the question but also during the middle of the generation, further showcasing its versatility.
	
	\section{Limitations}
	
	While this work focuses on improving the efficiency of LLM training through better sampling methods, there are two limitations. First, our approach lacks a strong theoretical guarantee, meaning that there is a possibility that future models, especially ones that are with different model architectures, may not benefit from it. Second, although our method is designed for training LLMs, the inference-time algorithm could potentially bypass safety measures by sampling out-of-distribution data. However, we argue that this concern can be inherently mitigated in models trained with our proposed sampling technique.
	
	\bibliography{main}
	
	\appendix
	
	\section{Implementation Details}
	
	In the paper, we proposed FIRE sampling, which is similar to CoT-decoding, and removed the need to calculate the confidence score. One of the biggest benefits of simplifying the method is getting an extremely easy implementation. For inference, we use vLLM \cite{kwon2023efficient} and do a two-stage sampling, with the first stage sampling only one token with high temperature and the second stage continuing the sampling with regular sampling. For training, we implement based on HybridFlow\cite{sheng2024hybridflow}, a newly released RLHF code base, which supports sampling with vLLM. Thus, we only changed the sampling part of the code in the RLHF framework. As shared in all experiments, the temperature used for the initial token is set at 30.
	
	In our experiment, we enumerate the parameters of top-p sampling, top-k sampling, and min-p sampling. We list all the parameters we have tried in the next section. Due to computation costs, some of the models are not enumerated in the same number as others. However, our conclusion that FIRE outperforms regular sampling is consistent, as we will show later.
	Specifically for MBPP(+), and for Qwen-RL, the model after our fine-tuning, we test on a single hyperparameter combination of $top-p=0.9$, $top-k=16$, which follows the best configuration from previous trials. For Qwen2.5-72b-Instruct, we follow the recommended hyperparameters of $top-p=0.8$, $top-k=16$. For reinforcement learning problems, we use the default parameters in HybridFlow, specifically, $top-k=16$, $top-p=1.0$. For training with FIRE sampling, to enable PPO to accept the relatively out-of-distribution samples, we change the clipping ratio for PPO from 0.2 to 0.5. We observe that for PPO+FIRE to use the original clip rate, it will generally match the original performance, while pure PPO with a higher clip ratio will lead the training to a failure and converge to a pass rate close to 0. 
	
	In the paper, we use three different datasets: GSM8K, MATH, and MBPP(+). GSM8K is a dataset with 8.5K total instances of math problem, of which 7.5K is in the training set and 1.3K is in the test set. MATH is a math dataset that is slightly more difficult and more comprehensive than GSM8K, with 7.5K training data and 5K test data. MBPP is a benchmark consisting of around 1,000 crowd-sourced Python programming problems, and MBPP+ is a benchmark that enlarges MBPP with some harder problems, reaching around 35K total test problems. While MBPP+ is still under regular update, we use version 0.1.0 in our paper. 
	
	For the final part of the experiment about generation in the middle sequence, we use a dataset that guarantees a certain number of sentences of prefixes to be correct. Here, the sentences are defined based on '.' in the answer. This dataset is generated on the training set of the MATH dataset, for which we first use Qwen2-7b-Instruct to sample 10 responses for each question. Then, for each response, we enumerate the sentences and sample 20 times using different numbers of sentences as the prefix. Thus, we obtained an approximation of the point at which the original samples became wrong. Specifically, if one response is wrong before the number of lines we enumerate in Table~\ref{table:middle_high_temp}, we use all the prefix up to the point that is still correct for that response, i.e., if for a specific sample, the correct sentences are less than 2, 3rd-line pass rate will be calculated in the same way as PRM-line.
	
	\section{Extra Experiment Results}
	
	We provide our full inference experiment table in Table~\ref{table:dseek_all_res}, Table~\ref{table:gemma_all_res}, and Table~\ref{table:qwen_all_res}. We observe that among all hyperparameter combinations, FIRE stably outperforms regular sampling, starting at Pass@10 to Pass@20 and Pass@40. In most settings, FIRE is superior to regular sampling at Pass@5, and for certain settings in the MATH dataset, FIRE could even show an advantage in Pass@1. 
	
	\begin{table*}[]
		\tiny
		\centering
		\begin{tabular}{@{}llllllllllll@{}}
			\toprule
			Dataset & top-p & top-k & min-p & Sampling & Pass@1 & Pass@5 & Pass@10 & Pass@20 & Pass@30 & Pass@40 & EA@40 \\ \midrule
			\multirow{16}{*}{GSM8K}        & 0.7 & 16 & 0.01 & Reg & 87.19 & 93.40 & 95.07 & 95.53 & 95.98 & 96.29 & 1.96 \\
			& 0.7 & 16 & 0.01 & FIRE & 85.75 & 94.62 & 96.21 & 97.42 & 97.95 & 98.18 & 2.58 \\
			& 0.7 & 16 & 0.0 & Reg & 87.19 & 93.40 & 95.07 & 95.53 & 95.98 & 96.29 & 1.96 \\
			& 0.7 & 16 & 0.0 & FIRE & 85.75 & 94.62 & 96.21 & 97.42 & 97.95 & 98.18 & 2.58 \\
			& 0.7 & 32 & 0.01 & Reg & 87.72 & 93.63 & 94.77 & 95.45 & 96.13 & 96.36 & 1.97 \\
			& 0.7 & 32 & 0.01 & FIRE & 85.06 & 95.07 & 96.74 & 97.73 & 98.18 & 98.26 & 2.64 \\
			& 0.7 & 32 & 0.0 & Reg & 86.58 & 93.40 & 94.84 & 96.06 & 96.36 & 96.82 & 1.97 \\
			& 0.7 & 32 & 0.0 & FIRE & 85.67 & 94.84 & 96.66 & 97.57 & 98.10 & 98.26 & 2.63 \\
			& 0.9 & 16 & 0.0 & Reg & 87.41 & 94.16 & 95.68 & 96.66 & 97.35 & 97.57 & 2.26 \\
			& 0.9 & 16 & 0.0 & FIRE & 84.46 & 95.83 & 96.89 & 97.88 & 98.26 & 98.71 & 2.76 \\
			& 0.9 & 32 & 0.01 & Reg & 86.05 & 94.31 & 95.83 & 96.74 & 97.04 & 97.27 & 2.28 \\
			& 0.9 & 32 & 0.01 & FIRE & 84.76 & 94.84 & 96.59 & 97.88 & 98.33 & 98.41 & 2.86 \\
			& 0.9 & 32 & 0.0 & Reg & 86.43 & 94.01 & 95.53 & 96.66 & 97.19 & 97.35 & 2.29 \\
			& 0.9 & 32 & 0.0 & FIRE & 84.76 & 94.84 & 96.59 & 97.88 & 98.33 & 98.41 & 2.86 \\ \midrule
			\multirow{16}{*}{MATH}       & 0.7 & 16 & 0.01 & Reg & 51.04 & 64.66 & 69.20 & 72.36 & 73.86 & 74.82 & 5.08 \\
			& 0.7 & 16 & 0.01 & FIRE & 49.68 & 64.94 & 70.16 & 73.52 & 75.58 & 76.68 & 6.33 \\
			& 0.7 & 16 & 0.0 & Reg & 51.04 & 64.56 & 68.56 & 71.84 & 73.44 & 74.62 & 5.07 \\
			& 0.7 & 16 & 0.0 & FIRE & 49.58 & 65.48 & 70.22 & 74.34 & 76.02 & 77.16 & 6.34 \\
			& 0.7 & 32 & 0.01 & Reg & 51.00 & 64.36 & 68.42 & 71.74 & 73.46 & 74.38 & 5.06 \\
			& 0.7 & 32 & 0.01 & FIRE & 49.08 & 65.64 & 70.22 & 73.92 & 76.10 & 77.06 & 6.90 \\
			& 0.7 & 32 & 0.0 & Reg & 51.00 & 64.36 & 68.42 & 71.74 & 73.46 & 74.38 & 5.06 \\
			& 0.7 & 32 & 0.0 & FIRE & 49.08 & 65.64 & 70.22 & 73.92 & 76.10 & 77.06 & 6.90 \\
			& 0.9 & 16 & 0.01 & Reg & 50.42 & 64.98 & 69.44 & 72.98 & 75.00 & 76.08 & 5.66 \\
			& 0.9 & 16 & 0.01 & FIRE & 48.96 & 65.34 & 70.36 & 74.12 & 76.16 & 77.64 & 7.29 \\
			& 0.9 & 16 & 0.0 & Reg & 50.82 & 65.36 & 69.62 & 73.12 & 75.06 & 76.16 & 5.64 \\
			& 0.9 & 16 & 0.0 & FIRE & 48.26 & 65.00 & 69.98 & 74.42 & 76.18 & 77.64 & 7.26 \\
			& 0.9 & 32 & 0.01 & Reg & 50.00 & 65.40 & 69.32 & 72.88 & 74.72 & 75.98 & 5.65 \\
			& 0.9 & 32 & 0.01 & FIRE & 47.66 & 65.48 & 70.48 & 74.58 & 76.86 & 78.16 & 7.90 \\
			& 0.9 & 32 & 0.0 & Reg & 50.00 & 65.40 & 69.32 & 72.88 & 74.72 & 75.98 & 5.65 \\
			& 0.9 & 32 & 0.0 & FIRE & 47.66 & 65.48 & 70.48 & 74.58 & 76.86 & 78.16 & 7.90 \\ \bottomrule
		\end{tabular}
		\caption{Deepseek-coder-v2-Instruct on different datasets with regular sampling (Reg) and FIRE (ours). We show the pass rate with different number of samples (Pass@n), and the effective answers (EA) of the total 40 samples.}
		\label{table:dseek_all_res}
	\end{table*}
	
	\begin{table*}[]
		\tiny
		\centering
		\begin{tabular}{@{}llllllllllll@{}}
			\toprule
			Dataset & top-p & top-k & min-p & Sampling & Pass@1 & Pass@5 & Pass@10 & Pass@20 & Pass@30 & Pass@40 & EA@40 \\ \midrule
			\multirow{16}{*}{MATH}  
			& 0.7 & 16 & 0.01 & Reg & 15.90 & 29.94 & 36.40 & 42.36 & 45.74 & 48.14 & 8.44 \\
			& 0.7 & 16 & 0.01 & FIRE & 17.20 & 32.28 & 39.22 & 45.30 & 48.52 & 51.18 & 9.82 \\
			& 0.7 & 16 & 0.0 & Reg & 15.90 & 29.94 & 36.40 & 42.36 & 45.74 & 48.14 & 8.44 \\
			& 0.7 & 16 & 0.0 & FIRE & 17.20 & 32.28 & 39.22 & 45.30 & 48.52 & 51.18 & 9.82 \\
			& 0.7 & 32 & 0.01 & Reg & 15.78 & 29.84 & 36.16 & 41.70 & 45.20 & 47.70 & 8.40 \\
			& 0.7 & 32 & 0.01 & FIRE & 16.68 & 32.40 & 38.80 & 45.32 & 48.90 & 51.26 & 9.76 \\
			& 0.7 & 32 & 0.0 & Reg & 15.78 & 29.84 & 36.16 & 41.70 & 45.20 & 47.70 & 8.40 \\
			& 0.7 & 32 & 0.0 & FIRE & 16.68 & 32.40 & 38.80 & 45.32 & 48.90 & 51.26 & 9.76 \\
			& 0.9 & 16 & 0.01 & Reg & 14.74 & 30.46 & 37.02 & 43.30 & 46.98 & 49.20 & 9.23\\
			& 0.9 & 16 & 0.01 & FIRE & 15.12 & 31.48 & 38.30  & 45.48 & 48.90 & 51.48 & 10.39 \\
			& 0.9 & 16 & 0.0 & Reg & 14.74 & 30.46 & 37.02 & 43.30 & 46.98 & 49.20 & 9.23 \\
			& 0.9 & 16 & 0.0 & FIRE & 15.12 & 31.48 & 38.30 & 45.48 & 48.90 & 51.48 & 10.39 \\
			& 0.9 & 32 & 0.01 & Reg & 14.58 & 30.16 & 36.20 & 42.28 & 45.98 & 48.34 & 9.17 \\
			& 0.9 & 32 & 0.01 & FIRE & 15.04 & 31.24 & 37.60 & 44.26 & 47.84 & 50.54 & 10.34 \\ 
			& 0.9 & 32 & 0.0 & Reg & 15.02 & 30.06 & 36.48 & 43.12 & 46.52 & 49.08 & 9.15 \\
			& 0.9 & 32 & 0.0 & FIRE & 14.58 & 31.40 & 38.36 & 44.92 & 48.48 & 51.06 & 10.35 \\\midrule 
			\multirow{8}{*}{GSM8K}& 0.7 & 16 & 0.01 & Reg & 36.54 & 66.41 & 75.66 & 82.34 & 84.46 & 86.81 & 3.86 \\
			& 0.7 & 16 & 0.01 & FIRE & 32.45 & 66.57 & 76.57 & 83.32 & 85.97 & 87.26 & 3.97 \\
			& 0.7 & 16 & 0.0 & Reg & 36.54 & 66.41 & 75.66 & 82.34 & 84.46 & 86.81 & 3.86 \\
			& 0.7 & 16 & 0.0 & FIRE & 32.45 & 66.57 & 76.57 & 83.32 & 85.97 & 87.26 & 3.97 \\
			& 0.7 & 32 & 0.01 & Reg & 36.92 & 67.25 & 75.66 & 82.11 & 84.08 & 85.52 & 3.91\\
			& 0.7 & 32 & 0.01 & FIRE & 31.24 & 66.79 & 76.27 & 82.87 & 85.97 & 87.57 & 4.01 \\
			& 0.7 & 32 & 0.0 & Reg & 36.92 & 67.25 & 75.66 & 82.11 & 84.08 & 85.52 & 3.91 \\
			& 0.7 & 32 & 0.0 & FIRE & 31.24 & 66.79 & 76.27 & 82.87 & 85.97 & 87.57 & 4.01 \\ \bottomrule
		\end{tabular}
		\caption{Gemma-2-2b-it on different datasets with regular sampling (Reg) and FIRE (ours). We show the pass rate with different number of samples (Pass@n), and the effective answers (EA) of the total 40 samples.}
		\label{table:gemma_all_res}
	\end{table*}
	
	\begin{table*}[]
		\tiny
		\centering
		\begin{tabular}{@{}llllllllllll@{}}
			\toprule
			Dataset & top-p & top-k & min-p & Sampling & Pass@1 & Pass@5 & Pass@10 & Pass@20 & Pass@30 & Pass@40 & EA@40 \\ \midrule
			\multirow{8}{*}{GSM8K}     & 0.7 & 16 & 0.01 & Reg & 66.72 & 89.23 & 92.80 & 94.47 & 95.07 & 95.83 & 2.61 \\
			& 0.7 & 16 & 0.01 & FIRE & 66.49 & 92.87 & 94.92 & 96.66 & 97.19 & 97.35 & 3.08 \\
			& 0.7 & 16 & 0.0 & Reg & 66.72 & 89.23 & 92.80 & 94.47 & 95.07 & 95.83 & 2.61 \\
			& 0.7 & 16 & 0.0 & FIRE & 66.49 & 92.87 & 94.92 & 96.66 & 97.19 & 97.35 & 3.08 \\
			& 0.7 & 32 & 0.0 & Reg & 67.02 & 89.16 & 92.27 & 94.31 & 95.30 & 95.91 & 2.58 \\
			& 0.7 & 32 & 0.0 & FIRE & 66.34 & 92.95 & 95.75 & 97.19 & 97.88 & 98.26 & 3.17 \\
			& 0.9 & 16 & 0.0 & Reg & 64.52 & 90.83 & 94.16 & 95.75 & 96.89 & 97.42 & 2.96 \\
			& 0.9 & 16 & 0.0 & FIRE & 64.22 & 92.27 & 95.07 & 97.04 & 97.65 & 97.95 & 3.33 \\    \midrule
			\multirow{16}{*}{MATH}    & 0.7 & 16 & 0.01 & Reg & 35.80 & 59.40 & 66.40 & 71.68 & 74.22 & 75.76 & 6.47 \\
			& 0.7 & 16 & 0.01 & FIRE & 40.26 & 63.74 & 69.98 & 75.08 & 77.42 & 78.90 & 7.86 \\
			& 0.7 & 16 & 0.0 & Reg & 35.80 & 59.40 & 66.40 & 71.68 & 74.22 & 75.76 & 6.47 \\
			& 0.7 & 16 & 0.0 & FIRE & 40.26 & 63.74 & 69.98 & 75.08 & 77.42 & 78.90 & 7.86 \\
			& 0.7 & 32 & 0.01 & Reg & 36.42 & 59.42 & 66.22 & 71.10 & 73.52 & 75.26 & 6.47 \\
			& 0.7 & 32 & 0.01 & FIRE & 39.52 & 63.54 & 70.10 & 75.06 & 77.42 & 78.92 & 8.11 \\
			& 0.7 & 32 & 0.0 & Reg & 36.42 & 59.42 & 66.22 & 71.10 & 73.52 & 75.26 & 6.47 \\
			& 0.7 & 32 & 0.0 & FIRE & 39.52 & 63.54 & 70.10 & 75.06 & 77.42 & 78.92 & 8.11 \\
			& 0.9 & 16 & 0.01 & Reg & 35.30 & 59.48 & 66.16 & 71.86 & 74.68 & 76.60 & 7.44 \\
			& 0.9 & 16 & 0.01 & FIRE & 38.70 & 62.44 & 69.50 & 74.64 & 77.36 & 78.86 & 8.76 \\
			& 0.9 & 16 & 0.0 & Reg & 35.30 & 59.48 & 66.16 & 71.86 & 74.68 & 76.60 & 7.44 \\
			& 0.9 & 16 & 0.0 & FIRE & 38.70 & 62.44 & 69.50 & 74.64 & 77.36 & 78.86 & 8.76 \\
			& 0.9 & 32 & 0.01 & Reg & 35.82 & 59.84 & 66.70 & 72.02 & 74.64 & 76.38 & 7.40 \\
			& 0.9 & 32 & 0.01 & FIRE & 37.44 & 62.72 & 69.50 & 75.08 & 77.50 & 79.08 & 9.03 \\
			& 0.9 & 32 & 0.0 & Reg & 35.14 & 59.84 & 66.80 & 72.34 & 74.72 & 76.40 & 7.43 \\
			& 0.9 & 32 & 0.0 & FIRE & 36.70 & 62.52 & 69.12 & 74.54 & 77.10 & 79.04 & 9.04 \\  \bottomrule
		\end{tabular}
		\caption{Qwen2-7B-Instruct on different datasets with regular sampling (Reg) and FIRE (ours). We show the pass rate with different number of samples (Pass@n), and the effective answers (EA) of the total 40 samples.}
		\label{table:qwen_all_res}
	\end{table*}
	
\end{document}